%% file: arXiv.tex
\documentclass[runningheads]{llncs}

\usepackage{booktabs} 
\usepackage{subfig}
\usepackage{graphicx}
\usepackage{colortbl}
\usepackage{algorithm}
\usepackage{algorithmic}
\usepackage{bm}
\usepackage{amsmath}
\usepackage{soul}
\usepackage{rotating}
\usepackage{multirow}
\usepackage{verbatim}
\usepackage[normalem]{ulem}
\usepackage{xcolor}

\usepackage{marvosym}
\usepackage{multiobjective}

\usepackage{ulem}
\usepackage{enumitem}
\setlist{leftmargin=4mm}

\usepackage[T1]{fontenc}

\usepackage{tabularx}

\newcounter{ALC@tempcntr}

\definecolor{mypink}{cmyk}{0, 0.7808, 0.4429, 0.1412}

\definecolor{blue-light}{RGB}{66, 191, 244}

\newcommand{\mar}[1]{\textcolor{black}{#1}}
\newcommand{\yaf}[1]{\textcolor{black}{#1}}
\newcommand{\ric}[1]{\textcolor{black}{#1}}
\newcommand{\myr}[1]{\textcolor{black}{#1}}

\newcommand{\ignore}[1]{}

\begin{document}

\normalem

\sloppy 

\title{Multi-layer local optima networks for the analysis of advanced local search-based algorithms}

\titlerunning{Multi-layer local optima networks}

\author{Marcella Scoczynski Ribeiro Martins\inst{1} \and
Mohamed El Yafrani\inst{2} \and
Myriam R. B. S. Delgado\inst{1} \and
Ricardo L\"{u}ders\inst{1}
}
\authorrunning{Martins et al.}
\institute{
Federal University of Technology Paran\'a (UTFPR), Brazil\\
\and
Operations Research group, Aalborg University, Denmark\\
}

\maketitle

\begin{abstract}
A Local Optima Network (LON) is a graph model that compresses the fitness landscape of a particular combinatorial optimization problem based on a specific neighborhood operator and a local search algorithm. Determining which and how landscape features affect the effectiveness of search algorithms is relevant for both predicting their performance and improving the design process. This paper proposes the concept of multi-layer LONs as well as a methodology to explore these models aiming at extracting metrics for fitness landscape analysis. Constructing such models, extracting and analyzing their metrics are the preliminary steps into the direction of extending the study on single neighborhood operator heuristics to more sophisticated ones that use multiple operators. Therefore, in the present paper we investigate a two-layer LON obtained from instances of a combinatorial problem using bit-flip and swap operators. First, we enumerate instances of NK-landscape model and use the hill climbing heuristic to build the corresponding LONs. Then, using LON metrics, we analyze how efficiently the search might be when combining both strategies. The experiments show promising results and demonstrate the ability of multi-layer LONs to provide useful information that could be used for in metaheuristics based on multiple operators such as Variable Neighborhood Search.
\end{abstract}


 \input{sect1}

 \input{sect2}
  \input{sect3}
  \input{sect4}
  \input{sect5}

\bibliographystyle{plain}
\bibliography{references}

\end{document}

%% file: sect1.tex
\section{Introduction}
\label{sec:sect1}

Local Optima Networks (LONs) can be seen as  high level models of discrete (combinatorial) fitness landscapes~\cite{ochoa2008study}. In the conception of a single layer non-weighted  LON, nodes are local optima of the addressed problem instance  and edges are search transitions based on a specific neighborhood operator.
 
These basic models are therefore achievable given a local search algorithm and a set of enumerable instances~\cite{ochoa2014local}. Recently, different LON models and metrics have been investigated. Weighted LONs are addressed in \cite{ochoa2010first}, while neutral landscapes are considered in \cite{ochoa2017mapping}. In \cite{hernando2017local}, the authors investigate  compressed LONs. A sampling procedure to extract LONs of larger instances is the focus of  \cite{verel2018sampling} while Fractal (self-similarity)
measures are used in \cite{thomson2018fractal}.

Although recently they have been  considered to continuous spaces  \cite{LonContProb2017,LonContProbGecco2019}, the majority of works on LONs reported in the literature are devoted to  discrete
search spaces and combinatorial optimization \cite{tomassini2008complex,daolio2010local,verel2011localNK,daolio2012local,daolio2013local, thomson2017comparing,ochoa2017mapping,hernando2017local,verel2018sampling,Yafrani2018FLA}. In this case,  LON models have been used to capture in detail the number and distribution of local optima in the search space.  Nevertheless they are highly dependent on the specific neighborhood operator being addressed, exploring their structures serves as the basis of many theoretical studies on combinatorial optimization. 

In this paper we consider the NK-landscape model~\cite{kauffman1993origins,tomassini2008complex,chicano2017optimizing} to provide combinatorial landscapes. The joint use of LON and NK-landscape models has attracted attention by researchers in Fitness Landscape Analysis (FLA)~\cite{ochoa2008study,verel2008connectivity,tomassini2008complex,ochoa2010first,verel2011localNK}. This is mainly because, by changing the parameters $N$ and $K$, it is easy to range from simple to highly complex landscapes and impose different levels of difficulty to the addressed search algorithm.

LONs features  are known to be of utmost importance for understanding the search difficulty of the corresponding landscape as they provide useful insights about the landscape of the search space (search difficulty) and what problem properties are most influential, particularly in terms of the local optima distribution. 

As discussed in \cite{VNS_survey2016}, besides other important issues, the search for an optimal (or near-optimal) solution in combinatorial optimization must take into account the following aspects:
(i)  a local optimum relative to one neighborhood structure is not necessarily a local optimum for another neighborhood structure;
(ii) a global optimum is a local optimum with respect to all neighborhood structures. According to \cite{VNS_survey2016}, 
the first property suggests using several neighborhoods if local optima found are of poor quality. 
The second property might be exploited by increasingly using complex moves in order to find local optima with respect to all neighborhood structures used. 

Considering all the issues previously discussed (particularly the two aspects pointed out by \cite{VNS_survey2016}); and aiming to create the basis for an expansion of FLA to search algorithms that use multiple operators, we propose in this paper the multi-layer local optima network. We analyze LON features extracted from three different landscapes: one obtained using only the bit flip move, another resulted from the 1-swap move and the last one obtained by combining both bit-flip and 1-swap in a two-layer network. The present paper proposes to explore multi-layer local optimal networks  aiming to evaluate the use of multiple neighborhood operators as occurs in advanced local metaheuristics like Variable Neighborhood Search (VNS).

Therefore, the main contributions of this paper are:
(i) Proposing and formalizing the concept of multi-layer optimal networks (MLLONs);
(ii)
Presenting a methodology for exploring MLLONs and extracting metrics for fitness landscape analysis;
(iii) Presenting an analysis of each individual layer in an MLLON as well as their comparison with the whole multi-layer network.

Although they are well explored in complex networks~\cite{barrat2004architecture,costa2007characterization,kivela2014multilayer}, as far as we know, multi-layer networks had never been considered before in the context of FLA for optimization problems.  
Therefore, we believe that the present paper contributes with both, FLA and optimization areas, by presenting the preliminary steps into the direction of more robust  theoretical analyses of  sophisticated metaheuristics.

This paper is organized as follows. Section~\ref{sec:sect2}  provides background information on LONs, the addressed problem (NK Landscape models),
standard graph and complex network metrics 
as well as a brief description of some related approaches. Then,  Section~\ref{sec:proposal} describes in detail the proposed approach. Section~\ref{sec:exper_results} presents the  set up for experiments and  obtained results. 
Finally, Section~\ref{sec:conclusion} concludes the paper.

%% file: sect2.tex
\section{Background}
\label{sec:sect2}

This section provides basic information necessary to understand LON models (for fitness landscape analysis) and the NK Landscape models (the addressed problem). The section also discusses some related approaches focusing on LON variants and different application problems.


\subsection{Local Optima Networks}
\label{sec:LON}

A local optima network (LON) is a graph model 
used to represent landscape of combinatorial problems~\cite{ochoa2008study}. 
The networks can be explored using a set of metrics to characterize their landscapes and problem difficulty \cite{verel2011localESCAPE}.

\begin{figure}[!h]
\centering
  \includegraphics[scale=0.3]{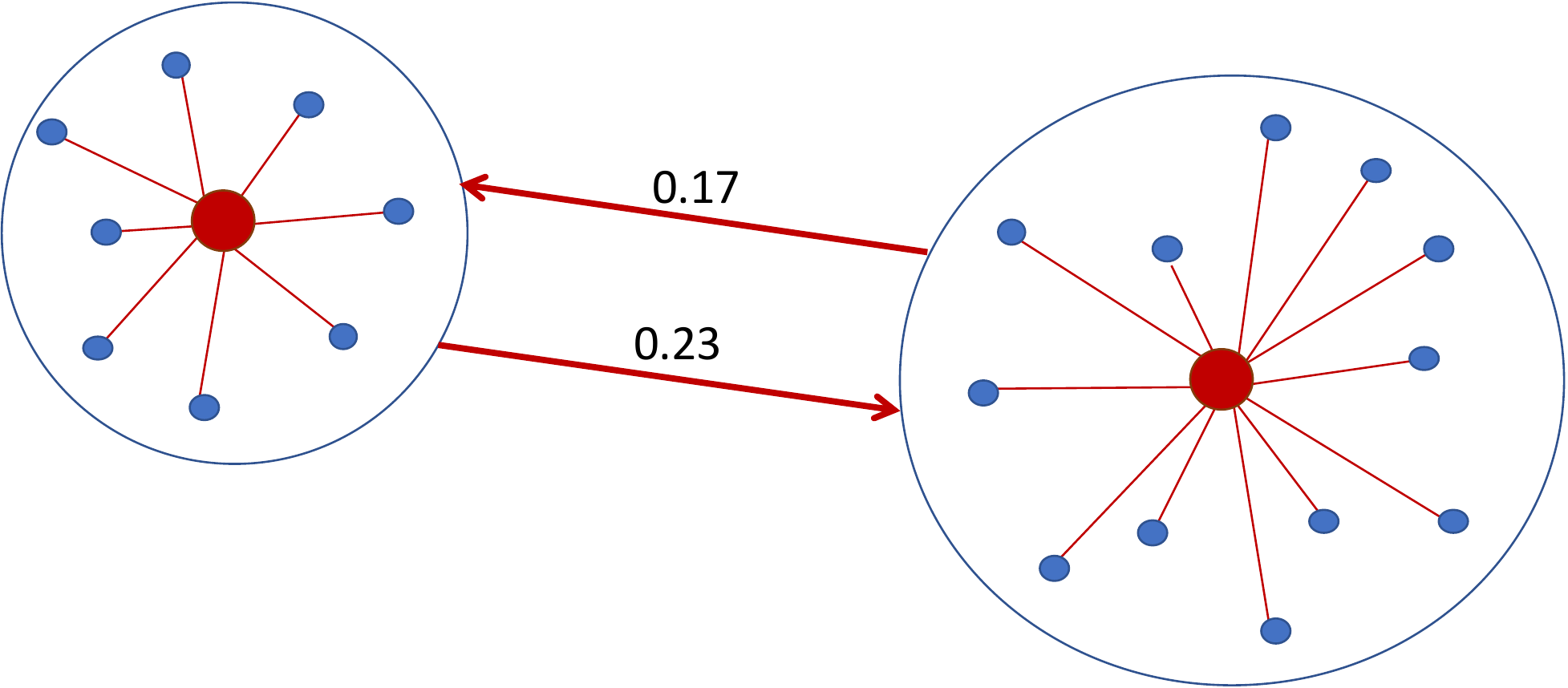}
  \caption{An example of a Local Optima Network with two basins connected by weighted edges}
   \label{fig:weighted_lon}
\end{figure}

In combinatorial optimization, the solutions of the addressed instance problem form a set that can be connected according to a given heuristic or move operators.  In a LON, the local search heuristic $\mathcal{H}$ maps a set of local optimum solutions $S^*$ from a solution space $S$~\cite{ochoa2008study}. 
  Considering a fitness function $F$, a solution $s^{*}$ in the solution space $S$ is a local maximum given a neighbourhood operator of type $t$ ($Nh_t$), iff $F(s^{*})\geq F(s), \forall s \in Nh_t(s)$.
  \myr{Figure  \ref{fig:weighted_lon} illustrates  a LON  with two subsets of solutions (small blue circles surrounded by large blue circles), from which two different local optima (red circles) can  be reached using a specific $\mathcal{H}$ and  a particular neighborhood operator. These subsets are called basins of attraction and can be connected by weighted or non-weighted edges.}
  
 \myr{To obtain the LON by enumerating all the solutions and then finding each basis of attraction with its associated local optima,  $\mathcal{H}$ starts from every solution $s$ in the solution space $S$, constructs the neighborhood structure of $s$  based on the neighborhood operator (bit-flip or swap for example) and using an update strategy (e.g., best improvement) selects one among the neighbors to be the next current solution. The process is repeated until a local optimum is reached.  For more details see Algorithm \ref{algo:hc} in Section \ref{sec:MLLONformalism}}

 \myr{The process of finding a local optima from $s$ involves constructing a neighborhood structure to move from $s$ to $s'$ during the search. The total number of solutions in the  neighborhood structure of a solution $s$ varies according to the  neighborhood operator considered. In this paper we address two different operators (bit-flip and swap) resulting in two different neighborhood structures. In the first case ($Nh_{\text{bf}}$),  bit-flip changes a bit of the binary solution from $0 \rightarrow 1$ and vice-versa. Therefore, for a binary string of size $N$,  there are $N$ neighbors for each solution. In the second case ($Nh_{\text{sw}}$),  the swap operator (more specifically the 1-swap operator) interchanges two randomly chosen bits of the string.
Therefore, the total of neighbors $T_N$ is calculated as the number of ones ($T_{1s}$) times number of zeros  present in $s$,  $T_N = T_{1s}(N-T_{1s})$.}

\myr{For each pair of solutions $s$ and $s'$, $p(s\rightarrow s')$ is the probability of  moving
from $s$ to $s'$ with the given neighborhood structure. Assuming the bit-flip neighborhood operator we have:
\[
p(s\rightarrow s')=
\left\{
	\begin{array}{ll}
	1/N & \mbox{if } s'\in Nh_{\text{bf}}(s) \\
	0 & \mbox{otherwise }
		\end{array}
\right.
\]
and for swap:
\[
p(s\rightarrow s')=
\left\{
	\begin{array}{ll}
	1/T_N & \mbox{if } s'\in Nh_{\text{sw}}(s) \\
	0 & \mbox{otherwise }
		\end{array}
\right.
\]}



As  previously discussed and depicted in Figure  \ref{fig:weighted_lon}, the basin of attraction of a local optimum $LO_j\in S$ is defined as $B_{j}=\{s\in S | \mathcal{H}(s)=LO_j\}$. This set, whose cardinality is given by $|B_{j}|$, contains all the solutions $s \in S $ from the local optimum $LO_j$ can be achieved when the local search heuristic $\mathcal{H}$ is applied. In Figure \ref{fig:weighted_lon}, two basins $B_1$ and $B_2$ are shown (large blue circles), each one composed of $|B_1|=8$ and $|B_2|=12$ solutions $s$ (small blue circles),  and their central nodes (red nodes) identify $LO_1$ and $LO_2$, respectively.
According to \cite{ochoa2014local}, the probability of moving from basin $B_i$ to $B_j$  is calculated as:
\begin{equation}
\label{eq:moveBetBasins}
    p(B_i\rightarrow B_j)=\frac{1}{|B_{i}|}\sum_{s\in B_i}\sum_{s'\in B_j}p(s\rightarrow s')
\end{equation}

The simplest LON model is built therefore by connecting  every two basins of attraction $B_i$ and $B_j$, whenever  $p(B_i\rightarrow B_j) > 0$. Assuming weighted networks \cite{barthelemy2005characterization} as a useful extensions of basic LONs,  weights  as those depicted in Figure \ref{fig:weighted_lon} ($w_{ij}= p(B_i\rightarrow B_j)$) can be also explored using specific FLA metrics \cite{ochoa2010first}.

Although  basins-transitions edges present a somewhat symmetric structure, when two nodes $i$ and $j$ are connected, both edges $e_{ij}$ and $e_{ji}$ are present 
\cite{verel2011localESCAPE}. 

\subsection{NK-landscape Models}
\label{sec:NKLandscape}

In this paper we consider the NK-landscape model~\cite{kauffman1993origins} as the application problem. These problems  have been designed to explore how the neighbourhood structure and the strength of interactions among neighbouring variables (subfunctions) are related to the search space ruggedness.

In the $NK$ model, $N$ refers to the number of (binary) variables, i.e. the string/solution
length, and $K$ to the strength of interaction, i.e. the number of variables that influence a particular variable. By increasing the value of $K$ from $0$ to $N - 1$, the landscapes can be tuned from smooth to rugged. 
The $K$ variables that contribute to the fitness of a string can be selected according to different models, and the two most widely studied ones
are the random neighborhood and the adjacent neighborhood models.
In this paper we adopt the general random model due to the fact that no
significant differences between the two were found 
in terms of the landscape
global properties \cite{kauffman1993origins}. 

\subsection{Related works}
\label{sec:relWorks}

Local optima networks have been explored on NK landscape problem \cite{tomassini2008complex,ochoa2008study}, including their extension to neutral fitness landscapes \cite{verel2011localNK}. Neutral networks consider solutions of equal (or quasi equal) objective function values.  Besides, several works analyzed the correlation between LON features and the performance of search heuristics in several problems~\cite{chicano2012local,daolio2012local,ochoa2014local}.

In \cite{ochoa2010first} the authors address  weighted LONs and consider first-improvement (greedy-ascent) hill-climbing algorithm, instead of a best-improvement (steepest-ascent) one, for the definition and extraction of the basins of attraction of landscape optima.  Results suggest differences with respect to both the network connectivity, and the nature of the basins of attraction.  
The paper discusses the impact of these structural differences between the two models (first versus best) in the behavior of search heuristics.

Permutation-based problems have also been subject to LON analyses~\cite{daolio2013local}. 
In \cite{ochoa2017mapping} the authors considered the LON model to characterize and visualize the global structure of travelling salesperson (TSP) fitness landscapes of different classes, including random and structured real-world instances of realistic size. 
The authors also investigated how to visualize neutral landscapes featured by the structured TSP instances.

Another LON variant is discussed in \cite{hernando2017local}. The Compressed Local Optima Network  is used to investigate different landscapes for the Permutation Flowshop Scheduling Problem (PFSP), exploring the network features to find differences between the landscape structures. The author analysed which features impact the performance of an iterated local search heuristic.

Recently, the work presented in \cite{Yafrani2018FLA} investigated two hill climbing local searches on a new combinatorial problem. The authors investigated the modeled LONs to explore and understand the difficulty of Travelling Thief Problem (TTP) instances.  
The results showed that certain operators can provide LONs with disconnected components and sometimes there exist exploitable correlations of node degree, basin size, and fitness.

The authors in \cite{verel2018sampling} addressed the limitation to fully enumerate all local optima, presenting a sampling procedure to extract LONs of larger instances and estimate their metrics without losing much in accuracy. They propose to increase the number of relevant features that could be used for performance prediction of Quadratic Assignment Problem (QAP) algorithms, in order to find whether LON features could be more correlated with performance than basic fitness landscape features. The experiments produced reliable results.

Another recent work in \cite{thomson2018fractal} investigated the use of fractal (self-similarity)
measures in local optima space in a fitness landscape. The authors applied  fractal dimension and associated metrics over NK landscape instances. The results showed a correlation between fractal geometry in the networks and the performance for search algorithms.





%% file: sect3.tex
\section{Proposed model}
\label{sec:proposal}

In this section we present the details of the proposed approach, including an overview on its basic concepts, the associated formalism and how to explore it through standard and new FLA metrics.

\subsection{Multi-layer Local Optima Networks}

\begin{figure}[!h]
\centering
  \includegraphics[scale=0.45]{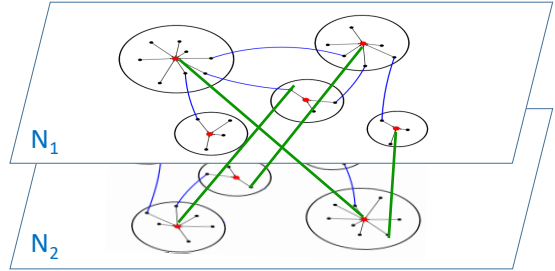}
  \caption{An example of a Multi-layer Local Optima Network composed of two layers: \myr{blue thin edges connect basins at the same layer (the intra-layer connections) while inter-layer connections are shown as red (connecting the same local optima) and green edges (connecting any two solutions)}.  The weights are omitted for simplification and readability purposes.
}
  \label{fig:MLON}
\end{figure}

As a main contribution of the present paper we propose the multi-layer Local Optima Network (MLLON).
Figure \ref{fig:MLON} illustrates an example of an MLLON composed of 
two layers $Nh_1$ and $Nh_2$, each one obtained using a different neighborhood operator.

 LON representations actually focus on local optima and edges only. However, in Figure  \ref{fig:MLON} we also shown the details of each basin of attraction to facilitate the explanation of the proposed model. In the proposed model, connections between basins located at the same layer (the intra-layer connections shown as blue thin edges in Figure \ref{fig:MLON}), represent the usual transitions between  pairs of basins as described in Section \ref{sec:LON}. The novelty of this paper is with respect to the possibility of {\bf{connecting  layers}} through {\bf{inter-layer}} connections. These connections allow one to {\bf{interchange}} the landscape analysis between  different {\bf{neighborhood operators}}. Inter-layer connections are shown as {\bf{red}} and {\bf{green}} edges and can be divided in two main groups: mirror and basin inter-layer connections. 

In the first case (mirror inter-layer connections), they 
appear as red edges and
indicate that the local optima are the same.  In other words, if a local optimum  appears in both layers, there is a red edge  connecting them, indicating the possibility of changing the neighborhood operator due to this common local optimum. In the second case (basin inter-layer connections), they 
appear as green edges and
 indicate that there is at least one common solution in both basins, which turns possible to the search  moving from   the other layer (due to the common solution) and using the new corresponding operator. Therefore, in MLLONs,  one can navigate from one  layer to another whenever a local optimum occurs in both layers or two basins in different layers share solutions. 


\ignore{
\begin{algorithm}[ht]
\caption{Generic VNS algorithm\label{algo:vns}}
\begin{algorithmic}[1]
\REPEAT 
    \STATE $k\gets1$ 
    \REPEAT
   \STATE $x' \gets LocalSearch(x,k)$
   \STATE $x \gets NeighbourhoodChange(x,x',k)$
   \STATE $x \gets PerturbationMove(x)$
   \STATE $k\gets k+1$
   \UNTIL {$k=kmax$}
\UNTIL{$t>tmax$}
\end{algorithmic}
\end{algorithm}

\yaf{The main motivation of creating such model is to have the ability to analyze advanced local-search-based metaheuristics. In particular, the model is designed to fit into the VNS framework as illustrated in algorithm~\ref{algo:vns}.
}
}

\subsection{The MLLON Formalism}
\label{sec:MLLONformalism}

\myr{When dealing with more than one neighborhood operator as occurs in many search techniques, a fundamental concept is when is the moment to change the operator. In the context of MLLONs,  this means changing the layer associated with one particular operator to another layer associated to a different operator. In this paper we consider two different circumstances when the operator can change: when the search achieves a local optimum shared with other layers (called mirror), or when the search achieves a solution that occurs in two or more layers. In the first case we must set the probability of changing the operator due to a mirror ($P_{co}^m$). In the second case we must set the probability of changing the operator due to a common solution in two different layers ($P_{co}^{\neq}$).}

Considering $p^M(B_i\rightarrow B_j)$ as the probability of moving from any basin $B_i$ to any basin $B_j$ in the multiple layer network, we can define a more general concept regarding the probability of transition between a pair of basins as:
\begin{equation}
p^M(B_i\rightarrow B_j)=
\left\{
	\begin{array}{cll}
		P_{sl}  & \mbox{if }L_i = L_j\mbox{,} &  \mbox{} LO_i = LO_j  \\
		P_{co}^m & \mbox{if } L_i \neq L_j\mbox{,}  &  \mbox{} LO_i = LO_j  \\
		          &       &   \\
	 p(B_i\rightarrow B_j) &   \mbox{if }L_i = L_j\mbox{,}  &  \mbox{} LO_i \neq LO_j \\
		  P_{co}^{\neq} \; p(B_i \bigcap  B_j) & \mbox{if } L_i \neq L_j\mbox{,} &  \mbox{} LO_i \neq LO_j  
	\end{array}
\right.
\label{eq:probMLON}
\end{equation}
 \myr{where  $P_{sl}$ is the probability of a  self-loop in a particular layer; $P_{co}^m \in (0,1]$ is the probability of changing the operator (co) when we find a mirror, i.e., when the local optima are the same ($LO_i = LO_j$) but located at different layers ($L_i \neq L_j$); $P_{co}^{\neq} \in [0,1]$ is the probability of changing the operator  when the local optima are different ($LO_i \neq LO_j$), and located at different layers  ($L_i \neq L_j$), what means that the two different basins of attraction $B_i$ and $B_j$ are also located at different layers.}
 
   In an MLLON,  when the basins  are in the same layer (i.e., $L_i = L_j$ for the pair $ij$), we have  the same probability of transition of a LON, what means $p^M(B_i\rightarrow B_j)=p(B_i\rightarrow B_j)$ with $p(B_i\rightarrow B_j)$ given by Equation \ref{eq:moveBetBasins}. 
   
   The novelty in this paper is the possibility of moving between layers due to the change in the neighborhood operator. In this case,  we only have to set  the value of parameter  $P_{co}^m$ when we have a mirror (same LO located at different layers), otherwise we set $P_{co}^{\neq}$ and calculate the probability of transition which is given by Equation  \label{eq:probBiIntBj_b}.
   
   
\begin{equation}
p(B_i \bigcap  B_j)= \frac{{|B_i \bigcap B_j|}}{{|B_i \bigcup B_j|}}
\label{eq:probBiIntBj_b}
\end{equation}
where $|B_i \bigcap B_j|$ is the cardinality of the set $B_i \bigcap B_j=\{s'|\;s' \in B_i \textbf{ and } s' \in B_j\}$ and $|B_i \bigcup B_j|$ is the cardinality of the set $B_i \bigcup B_j=\{s|\;s \in B_i\textbf{ or } s \in B_j\}$.

Based on the previous discussion, the probability of transition between two different basins  can be summarized as shown in Table~\ref{tab:MLONprob}.
\begin{table}[!h]
\small
\caption{Probability of transitions between pair of Basins in MLLONs ($p^M(B_i\rightarrow B_j)$)  color-coded based on Figure~\ref{fig:MLON}.}
\label{tab:MLONprob}
\begin{tabular}{ccccc}
\cline{1-3}
\multicolumn{1}{l|}{Type}           & \multicolumn{1}{c|}{Same  $LO$} & \multicolumn{1}{c}{Different  $LO$} &  &  \\ \cline{1-3}
\multicolumn{1}{l|}{Intra-layer} & \multicolumn{1}{l|}{$\color{gray} p^M(B_i\rightarrow B_j)=P_{sl}$}      & \multicolumn{1}{l}{$\color{blue}p^M(B_i\rightarrow B_j)= p(B_i\rightarrow B_j)$}     &  &  \\ \cline{1-3}
\multicolumn{1}{l|}{Inter-layer} & \multicolumn{1}{l|}{$\color{red} p^M(B_i\rightarrow B_j)=P_{co}^m$}      & \multicolumn{1}{l}{$\color{green} p^M(B_i\rightarrow B_j)=P_{co}^{\neq} \; p(B_i \bigcap B_j)$}     &  &  \\ \cline{1-3}
\end{tabular}
\normalsize
\end{table}

As depicted in Figure  \ref{fig:MLON} and detailed in Table \ref{tab:MLONprob},  MLLONs encompass  different connections between each pair of basins: intra and inter-layers. It is important to point out that edges between basins (blue, red and green) can identify different kinds of connections, whose weights (probabilities) are calculated based on different sources of information (as shown in each cell of Table \ref{tab:MLONprob}). However, as discussed in the previous section, for sake of simplicity, neither weight values nor  directions are presented in Figure  \ref{fig:MLON}.

According to the code colors linking  Table \ref{tab:MLONprob}  and Figure  \ref{fig:MLON}, the blue  cell in Table \ref{tab:MLONprob} has a correspondence with blue edges in Figure  \ref{fig:MLON} and identifies the {\bf{intra-layer}} connections linking two basins whose local optima are different. \myr{Notice that in Figure  \ref{fig:MLON}, {\bf{intra-layer edges}} connecting two basins with the same local optima are impossible to occur. This is because although the formalism allows the occurrence of self-loops with probability $P_{sl}$, in the  model addressed in the experiments there is no duplicated local optimum in the same layer and no loop is allowed. Therefore, the corresponding cell is grey and the associated probability is set as 0 in the experiments.} However, assuming a looping edge as possible, this information could also be considered by the model. The red  cell has a correspondence with  red edges and identifies the {\bf{inter-layer}} connections linking two basins with the same local optima \myr{(notice that same local optima does not mean the same basins of attraction, since their solutions can be different)}. The green  cell has a correspondence with green edges and identifies the {\bf{inter-layer}} connections linking two basins with different local optima. 

As can be seen in the first line of Table \ref{tab:MLONprob}, LONs are a special case of MLLOs. \myr{Therefore, in the case of a unique layer (LON), $L_i = L_j$ for all the basins and  $P_{co}^m=P_{co}^{\neq}=0$, since there is no mirror or other operator to be considered. Moreover, $P_{sl} > 0$  indicates the possibility of self-loops in the LON,    $p^M(B_i\rightarrow B_j)=p(B_i\rightarrow B_j) $ and can be calculated as Equation \ref{eq:moveBetBasins}}. 
 
A particular MNLON model can be achieved considering $P_{co}^{m}=1$ and $P_{co}^{\neq}=0$, when the change in the neighborhood operator occurs only when the search achieves a mirror (i.e., the change occurs only when the search achieves a shared local optimum occurring in both layers).

In the experiments we consider this case, and  assume that layer $Nh_1$ is based on bit-flip  and  layer  $Nh_2$ is based on 1-swap,
both built from  NL-landscape models (for $N=18$ and $K$ ranging from $2$ to $16$). 


 Aiming to build each layer, the Best-improvement - Hill Climbing algorithm, presented in Algorithm \ref{algo:hc}, is applied to determine the local optima and therefore define the
basins of attraction.  The entire neighborhood is explored and the best solution is returned as the $LO$.  

\begin{algorithm}[ht]
\small
\caption{Algorithm based on Best-improvement - HC for constructing the Basins of attraction\label{algo:hc}}
\begin{algorithmic}[1]

\REQUIRE The complete set of solutions $S$ and the type $t$ of the neighborhood $Nh_t(.)$
\REPEAT
\STATE Initial solution $s_0\in S$
\STATE $s \gets s_0$
\REPEAT 
    \STATE Choose $s'\in Nh_t(s)$ such that $F(s')=max_{x\in Nh_t(x)}F(x)$  
   \IF{$F(s)<F(s')$}
   \STATE $s\gets s'$
   \ENDIF
\UNTIL{$s$ is a local optima}
\STATE $LO = s$, $s_0$ is part of its basin of attraction
\UNTIL{all solutions in $S$ have been visited as initial solutions $s_0$ }
\ENSURE 
\end{algorithmic}
\end{algorithm}

\subsection{Exploring MLLONs by means of graph theory metrics}

In the calculation of MLLON metrics, the multi-layer network can be transformed into a single-layer one using different methods. Aggregation and  flattening are the most popular ones~\cite{dickison2016multilayer}. \ric{In an aggregated network, nodes from different layers are aggregated in a single node by using different methods while flattening preserve connected nodes from different layers by adding an edge between them.} In  this paper we flattened the multi-layer network, then LON metrics were  straightforwardly applied.

In this paper, we consider four types of metrics to explore LONs and  MLLONs:
\begin{enumerate}
    \item descriptive statistics: 
    number of nodes (\textbf{$n_v$}),   
     number of edges (\textbf{$n_e$}),
    weighted assortativity or affinity (\textbf{$knn$}) and the fitness-fitness correlation  (\textbf{$fnn$});

    \item local metrics: cumulative degree distribution, correlation between degree and basin of attraction size, and correlation between fitness of local optima and basin of attraction size;
    \item global metrics: average weighted clustering coefficient (\textbf{$\overline{wcc}$}), 
    average weighted clustering coefficient for a random graph (\textbf{$\overline{wcc_r}$}), 
    average shortest path length (\textbf{$\overline{l}$}), 
    strength (\textbf{$\overline{st}$}), disparity 
    (\textbf{$\overline{y_2}$}), degree of outgoing edges
    (\textbf{$\overline{z_{out}}$}) and the average shortest path to global optima    (\textbf{$\overline{l_{GO}}$ }). Apart from (\textbf{$\overline{l_{GO}}$ }), that is new in the context of FLA, and has been explored for the first time in this paper, these metrics are significantly related to search performance and have been already considered in other works\cite{chicano2012local, daolio2012local}.
\end{enumerate}


As performed in \cite{Yafrani2018FLA}, in terms of descriptive statistics,  we start from the simplest ones, number of nodes  and  edges ($\textbf{$n_v$}$  and \textbf{$n_e$}, respectively) to analyze the impact of each neighborhood operator in the (ML)LON size. Aiming to evaluate the  neighborhood connectivity behavior  we consider $knn$, as it measures the nearest-neighbors degree correlation. In this case,  the  network is said to show assortative mixing if  high degree nodes  are supposed to be connected to other high degree nodes \cite{newman2002assortative}.  The Pearson correlation coefficient is applied to measure the network assortativity;  positive values indicate similarity in pair degrees and negative values represent relationships between nodes of different degree. 
As an attempt to evaluate the multi-modality degree of the landscape, we consider as the last descriptive statistic the $fnn$ which represents the fitness-fitness correlation and measures the correlation between the fitness values of adjacent local optima. As  in the case of $knn$, it also applies the Pearson correlation coefficient between the fitness value of a node and the weighted-average of its nearest neighbours fitness \cite{verel2018sampling}.

In terms of local metrics, we study the cumulative degree distribution, the correlation between the strength and basin size, and the correlation between fitness and basin size, since analysing the behavior of these distributions is useful to understand complex network structures \cite{antoniou2008statistical, Yafrani2018FLA, jakobovic2019characterisation}.

In the context of global features, i.e. those calculated considering the whole network, we consider 8 metrics. We start with   the average weighted clustering coefficients (\textbf{$\overline{wcc}$}), that 
measures cliquishness of a neighborhood \cite{barrat2004architecture}.
$\overline{wcc_r}$ is the average clustering coefficients of corresponding random graphs (i.e. random graphs with the same number of vertices and mean degree).
The average shortest path lengths between any two local optima is defined as $\overline{l}$ . The strength $\overline{st}$  for outgoing edges
measures the network weighted connectivity, while $\overline{y_2}$ is the average disparity for outgoing edges
which gauges the heterogeneity of
the contributions of the edges of each node to the total weight \cite{barthelemy2005characterization}.
$\overline{z_{out}}$ is the average out degree, i.e, the total number of outgoing edges. 
Besides these usual metrics, we consider the average shortest path to global optima    (\textbf{$\overline{l_{GO}}$}), measured according to the neighborhood operator applied to generate the corresponding LON.




%% file: sect4.tex
\section{Experiments and Results}
\label{sec:exper_results}

In the experiments conducted here we address the NK-landscape model~\cite{kauffman1993origins}, 
with  $N = 18$ variables, $K \in \{2, 4, 6, 8, 10, 12, 14, 16\}$ and  uniformly distributed sub-functions. This combination of parameters allows computing optimal solutions by exhaustively enumerating the solution space using reasonable computational resources. 


This paper aims to investigate  how efficiently is combining neighborhood moving operators (which can be accomplished by moving between multiple landscapes) when compared with each operator used in a stand-alone mode (ie. moving on one specific layer only). 

In the experiments, we generate therefore a multi-layer network composed of two layers corresponding  to the LONs using bit-flip and 1-swap operators respectively. 
To navigate in this multi-layer network,  edges inter-connecting two layers whose basins have the same local optima  have weight values set as $w_{ij}=p^M(B_i\rightarrow B_j)=P_{co}^m=1$. This assumption,  imposes no difficulty to jump from one layer to another, whenever the same local optima is identified in both layers.
Intra-layer edges, on the other hand, present weights according to their neighbourhood type (e.g. bit-flip for layer $N_1$ and 1-swap for layer $N_2$), whose values are defined as $w_{ij}=p(B_i\rightarrow B_j)$, with $p(B_i\rightarrow B_j)$ given by Equation \ref{eq:moveBetBasins}. Therefore,  low values indicate the difficulty to navigate through basin pairs.  The exception occurs in the calculation of the shortest path to the global optimum ($\overline{l_{GO}}$). This metric requires  complement values since lower values indicate better paths, then we consider $w_{ij}=1 - p(B_i\rightarrow B_j)$. 

Moreover, particularly for the multi-layer network we store the weighted edges, nodes and the corresponding layers as a supra-adjacent matrix \cite{de2013mathematical}. Therefore the multi-layer network can be flattened, preserving more information \cite{dickison2016multilayer}, and standard and complex metrics can be  applied. 

The networks are implemented and analyzed using Py3Plex \cite{Skrlj2019} and NetworkX \cite{hagberg2008exploring} from Python. Dijkstra's algorithm is applied to calculate the shortest path length. 

\subsection{Descriptive Statistics}

\begin{table}[htbp]
\centering
\caption{Descriptive statistics
}
\label{tab:statistics}
\begin{tabular*}{.64\textwidth}{l @{\extracolsep{\fill}} lllll}
\hline
\toprule
\textbf{Network type}& 
\textbf{$K$}& 
\textbf{$n_v$} & 
\textbf{$n_e$} & 
\textbf{$knn$}  & 
\textbf{$fnn$}\\ \hline
\multirow{8}{*}{Bit-flip} 
&$ 	2	$&$	32	$&$	370	$&$	-0.019	$&$	-0.356	$\\
&$	4	$&$	240	$&$	22280	$&$	-0.106	$&$	-0.604	$\\
&$	6	$&$	1080	$&$	199648	$&$	-0.122	$&$	-0.603	$\\
&$	8	$&$	2035	$&$	409390	$&$	-0.105	$&$	-0.649	$\\
&$	10	$&$	3593	$&$	687398	$&$	-0.075	$&$	-0.632	$\\
&$	12	$&$	5592	$&$	935578	$&$	-0.051	$&$	-0.616	$\\
&$	14	$&$	8313	$&$	1165660	$&$	-0.042	$&$	-0.587	$\\
&$	16	$&$	11383	$&$	1343544	$&$	-0.036	$&$	-0.569	$\\\hline
\multirow{8}{*}{Swap} 
&$ 	2	$&$	19317	$&$	2432458	$&$	-0.023	$&$	-0.562	$\\
&$	4	$&$	19894	$&$	2442764	$&$	-0.058	$&$	-0.521	$\\
&$	6	$&$	20720	$&$	2499472	$&$	-0.064	$&$	-0.584	$\\
&$	8	$&$	22046	$&$	2589536	$&$	-0.061	$&$	-0.603	$\\
&$	10	$&$	22906	$&$	2622780	$&$	-0.059	$&$	-0.617	$\\
&$	12	$&$	23489	$&$	2618234	$&$	-0.063	$&$	-0.612	$\\
&$	14	$&$	23734	$&$	2628042	$&$	-0.059	$&$	-0.611	$\\
&$	16	$&$	24441	$&$	2653482	$&$	-0.059	$&$	-0.604	$\\\hline
\multirow{8}{*}{Multi-layer}
&$ 	2	$&$	19349	$&$	2432888	$&$	-0.023	$&$	0.006	$\\
&$	4	$&$	20134	$&$	2465462	$&$	-0.051	$&$	0.006	$\\
&$	6	$&$	21800	$&$	2700928	$&$	-0.054	$&$	0.006	$\\
&$	8	$&$	24081	$&$	3002288	$&$	-0.058	$&$	0.005	$\\
&$	10	$&$	26499	$&$	3316056	$&$	-0.049	$&$	0.005	$\\
&$	12	$&$	29081	$&$	3562612	$&$	-0.023	$&$	0.004	$\\
&$	14	$&$	32047	$&$	3806266	$&$	-0.031	$&$	0.004	$\\
&$	16	$&$	35824	$&$	4013332	$&$	-0.084	$&$	0.003	$\\
\hline
\end{tabular*}
\end{table}

Table \ref{tab:statistics} presents descriptive statistics for a particular landscape.
We can observe that when $K$ increases, the number of local optima ($n_v$) and edges connections ($n_e$) increases accordingly. The increase in $n_v$ and $n_e$ is more significant for the bit-flip than the swap operator. Additionally, the value of these metrics is always higher for the swap operator. This shows that the search difficulty naturally increases when the ruggedness of the instance increases. However, the search difficulty using swap, although high, remains stable regardless of the ruggedness of the instances. 

Furthermore, there is a weak negative correlation for $knn$ for all values of $K$ (specially for the bit-flip), indicating smooth relationships between nodes of different degrees for the three cases. In other words, part of local optima with a large number of connections are connected to local optima with a low number of connections, which might provide a smooth navigation of the overall network \cite{daolio2012local}. 

The metric ${fnn}$ is particularly relevant for algorithms like VNS where fitness is applied as the acceptance criteria, as it measures correlation between the fitness values of adjacent local optima. In general, it is expected that a positive correlation contributes to the search process \cite{daolio2012local}. For both bit-flip and swap, the $fnn$ values are negative, indicating low connectivity between local optima with similar fitness value. However, this negative correlation becomes weak positive when combining the two landscapes using the multi-layer network. This can be explained by the addition of connections between two landscapes by linking local optima of same fitness with high weight values. We believe that using an aggregation method instead of flattening the multi-layer network could mitigate this effect, 
however, comparing these methods is out of the scope of this paper.

\subsection{Global Metrics}

\begin{table*}[htbp]
\centering
\caption{Global Metrics
\label{tab:global-metrics}
}
\begin{tabular*}{\textwidth}{l @{\extracolsep{\fill}} llllllll}
\hline
\toprule
\textbf{LON type}& 
\textbf{$K$}& 
\textbf{$\overline{wcc}$} & \textbf{$\overline{wcc_r}$} & 
\textbf{$\overline{l}$} & 
\textbf{$\overline{st}$} & 
\textbf{$\overline{y_2}$} &
\textbf{$\overline{z_{out}}$} &
\textbf{$\overline{l_{GO}}$} 
\\ \hline
\multirow{2}{*}{bit-flip} &$ 	2	$&$	0.524	$&$	0.752	$&$	1.67	$&$	11.12	$&$	0.09	$&$	11.56	$&$	1.13	$\\
&$	4	$&$	0.641	$&$	0.777	$&$	1.61	$&$	92.17	$&$	0.01	$&$	92.83	$&$	1.07	$\\
&$	6	$&$	0.489	$&$	0.343	$&$	1.83	$&$	184.08	$&$	0.01	$&$	184.86	$&$	1.41	$\\
&$	8	$&$	0.406	$&$	0.198	$&$	1.92	$&$	200.36	$&$	0.01	$&$	201.17	$&$	1.53	$\\
&$	10	$&$	0.317	$&$	0.107	$&$	2.02	$&$	190.47	$&$	0.01	$&$	191.32	$&$	1.80	$\\
&$	12	$&$	0.256	$&$	0.060	$&$	2.17	$&$	166.44	$&$	0.01	$&$	167.31	$&$	1.89	$\\
&$	14	$&$	0.210	$&$	0.034	$&$	2.37	$&$	139.34	$&$	0.01	$&$	140.22	$&$	2.15	$\\
&$	16	$&$	0.181	$&$	0.021	$&$	2.55	$&$	117.15	$&$	0.02	$&$	118.03	$&$	2.46	$\\
\midrule
\multirow{2}{*}{swap} &$ 	2	$&$	0.000	$&$	0.013	$&$	3.41	$&$	124.96	$&$	0.03	$&$	125.92	$&$	3.71	$\\
&$	4	$&$	0.000	$&$	0.012	$&$	3.41	$&$	121.84	$&$	0.03	$&$	122.79	$&$	3.10	$\\
&$	6	$&$	0.000	$&$	0.012	$&$	3.42	$&$	119.69	$&$	0.03	$&$	120.63	$&$	2.90	$\\
&$	8	$&$	0.000	$&$	0.011	$&$	3.43	$&$	116.53	$&$	0.03	$&$	117.46	$&$	3.75	$\\
&$	10	$&$	0.000	$&$	0.010	$&$	3.44	$&$	113.58	$&$	0.03	$&$	114.50	$&$	3.73	$\\
&$	12	$&$	0.000	$&$	0.009	$&$	3.46	$&$	110.55	$&$	0.03	$&$	111.47	$&$	3.10	$\\
&$	14	$&$	0.000	$&$	0.009	$&$	3.46	$&$	109.82	$&$	0.04	$&$	110.73	$&$	3.01	$\\
&$	16	$&$	0.000	$&$	0.009	$&$	3.47	$&$	107.66	$&$	0.04	$&$	108.57	$&$	3.44	$\\
\midrule
\multirow{2}{*}{multi-layer} &$ 	2	$&$	0.001	$&$	0.013	$&$	3.39	$&$	124.65	$&$	0.03	$&$	125.74	$&$	1.12	$\\
&$	4	$&$	0.007	$&$	0.012	$&$	3.33	$&$	121.37	$&$	0.03	$&$	122.45	$&$	1.06	$\\
&$	6	$&$	0.024	$&$	0.011	$&$	3.31	$&$	122.77	$&$	0.03	$&$	123.90	$&$	1.39	$\\
&$	8	$&$	0.034	$&$	0.010	$&$	3.30	$&$	123.51	$&$	0.03	$&$	124.67	$&$	1.53	$\\
&$	10	$&$	0.042	$&$	0.009	$&$	3.30	$&$	123.90	$&$	0.03	$&$	125.14	$&$	1.80	$\\
&$	12	$&$	0.048	$&$	0.008	$&$	3.31	$&$	121.20	$&$	0.03	$&$	122.51	$&$	1.88	$\\
&$	14	$&$	0.054	$&$	0.007	$&$	3.31	$&$	117.38	$&$	0.03	$&$	118.77	$&$	2.15	$\\
&$	16	$&$	0.056	$&$	0.006	$&$	3.33	$&$	110.57	$&$	0.03	$&$	112.03	$&$	2.31	$\\
\hline							
\end{tabular*}
\end{table*}

Table \ref{tab:global-metrics} summarizes the results for global metrics averaged for one landscape.
For the bit-flip operator, the average shortest path length is small and  proportional to $log(n_v)$ 
with mostly high clustering coefficients compared to an equivalent random graph. Therefore, the resulting graph is compatible with small-world networks.

On the other hand, 
despite its proportionality to $log(n_v)$,
$\overline{l}$ 
is larger for the swap operator than the corresponding values for the bit-flip with significantly small clustering coefficients. This typically indicates that the resulting graph does not fit properly in a small-world network model, yielding to a difficult in the network navigation.
This phenomenon naturally extends to the multi-layer network, but it is slightly alleviated thanks to the combination with the bit-flip-based LON.
The small clustering coefficient can be explained by the large number of local optima (number of nodes in Table~\ref{tab:statistics}) which yields to small basins and thus a small probability of neighborhood connections. In fact, in contrast to the bit-flip operator which gives a more connected network, the swap operator add the constraint of needing two unequal decision variable to apply the move, resulting in a possible not connected graph.
It is worth noting that similar results were obtained for the swap operator in some graphs when applied to the S-BOX problem~\cite{jakobovic2019characterisation}.

The metric $\overline{l_{GO}}$ shows that the bit-flip operator ensures reaching a global optimum in a small number of steps. While the cost of finding a global optimum significantly increases in the swap-based LON. This is consistent with the $\overline{l}$ results. Interestingly, combining both landscapes results in a network requiring less steps to find a global optimum. 
\myr{This means that, although using only bit-flip might be more efficient than using swap all the time, the search seems to benefit from the combination of both as swap may  provide  'jumps' that are not available when using only bit-flip.}   


The metric ${z_{out}^i}$ counts the number of outgoing edges of node $i$ (i.e., a local optimum $LO$ in the respective LON). According to \cite{daolio2012local} it is relevant to know whether all transitions have the same weight, or if there is a preferred direction. The disparity  $y_2^i = \sum_{j \neq i}(\frac{w_{ij}}{st_i})$ of node $i$ indicates the weight heterogeneity of outgoing edges \cite{barthelemy2005characterization}. It is related to the strength $st^i$: when all edges leaving a node $i$ have the same probabilities, the disparity is the inverse of
the out-degree $1/z_{out}^{i}$. Therefore, a low disparity indicates equally likely transitions, with no guidance in the search trajectory. This behavior seems to be present in the three network models since similar $\overline{y_2}$ 
values were found independently of ruggedness.


\subsection{Local Metrics}

\begin{figure*}[htbp]
\vspace{-0.5cm}
\centering
\rotatebox{90}{\hspace{11mm}\small{\textit{bit-flip}}}\rotatebox{90}{\rule{35mm}{0.3pt}}%
\quad 
\subfloat[Cumulative strength distribution]{
  \includegraphics[scale=0.2]{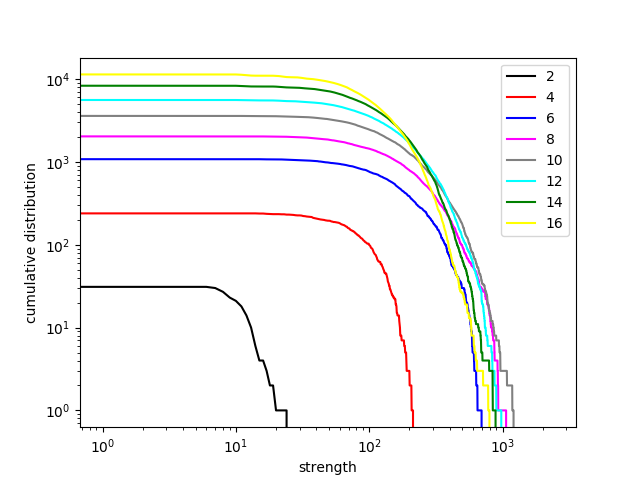}
  \label{fig:degreeDistbitflip}
}
\quad 
\subfloat[Strength vs. Basin of attraction size]{
  \includegraphics[scale=0.2]{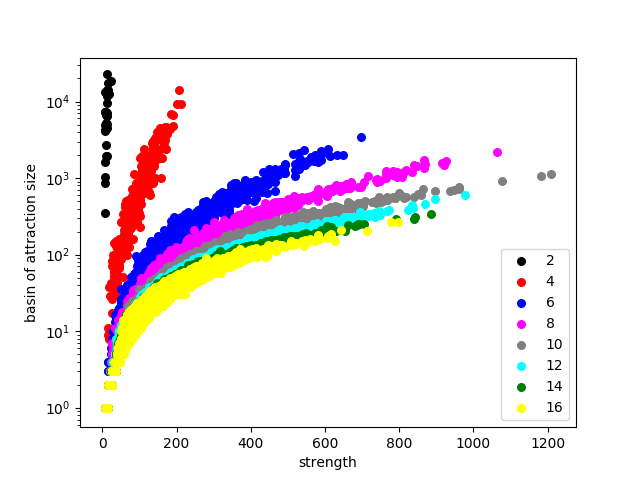}
 \label{fig:degreeBasinbitflip}
}
\quad 
\subfloat[Fitness vs. Basin of attraction size]{
  \includegraphics[scale=0.2]{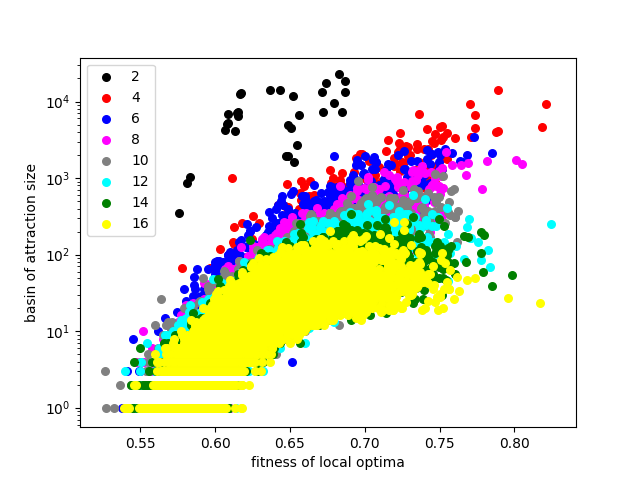}
  \label{fig:LOdegreeBasinbitflip}
}
\quad 
\rotatebox{90}{\hspace{11mm}\small{\textit{swap}}}\rotatebox{90}{\rule{35mm}{0.3pt}}%
\quad 
\subfloat[Cumulative strength distribution]{
  \includegraphics[scale=0.2]{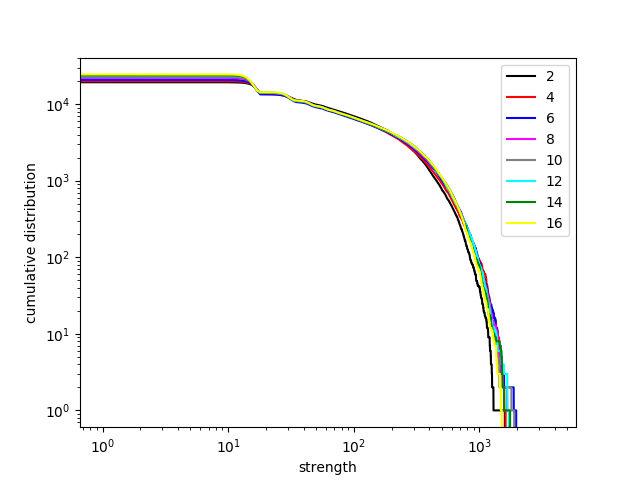}
  \label{fig:degreeDistswap}
}
\quad 
\subfloat[Strength vs. Basin of attraction size]{
  \includegraphics[scale=0.2]{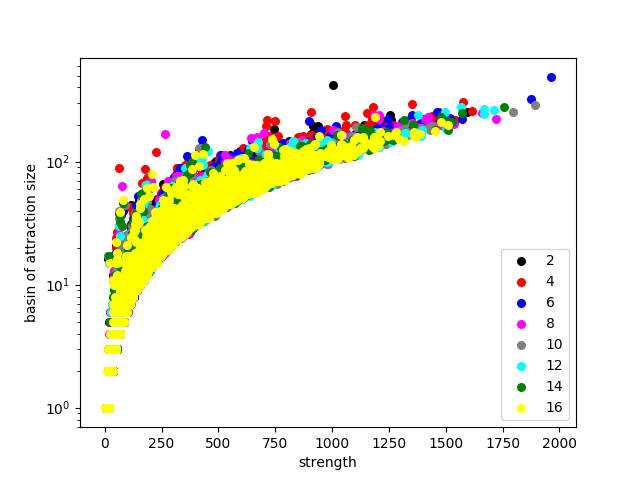}
  \label{fig:degreeBasinswap}
}
\quad 
\subfloat[Fitness vs. Basin of attraction size]{
  \includegraphics[scale=0.2]{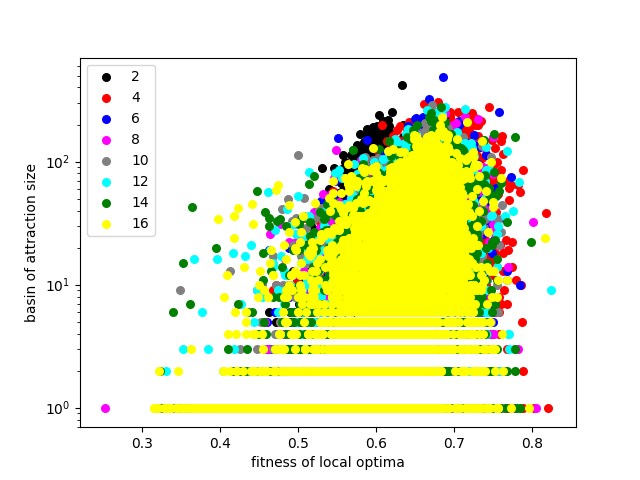}
  \label{fig:LodegreeBasinswap}
}
\quad 
\rotatebox{90}{\hspace{11mm}\small{\textit{multi-layer}}}\rotatebox{90}{\rule{35mm}{0.3pt}}%
\quad 
\subfloat[Cumulative strength distribution]{
  \includegraphics[scale=0.2]{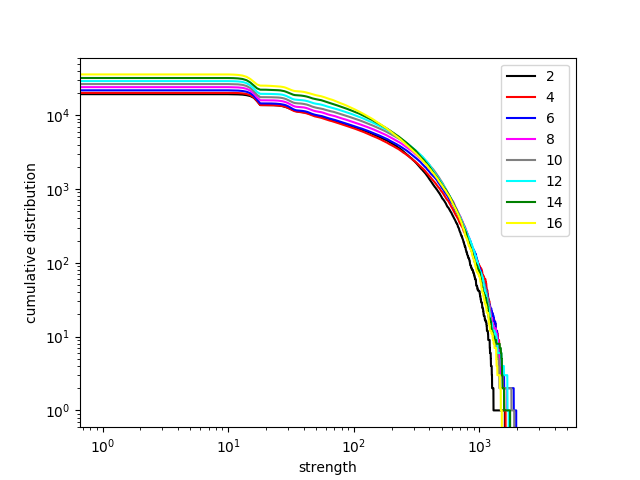}
  \label{fig:degreeDistmulti}
}
\quad 
\subfloat[Strength vs. Basin of attraction size]{
  \includegraphics[scale=0.2]{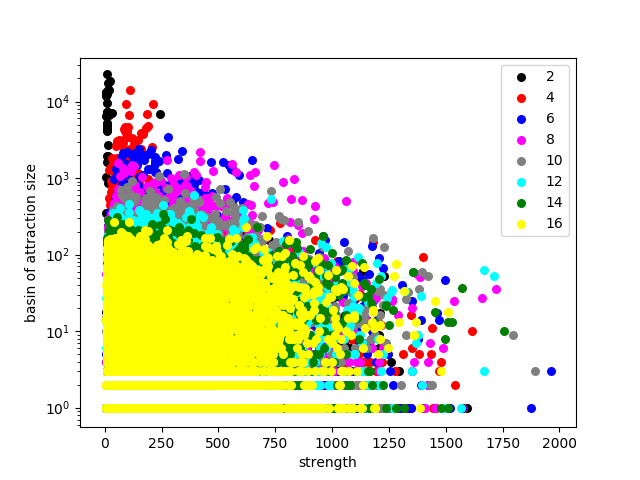}
  \label{fig:degreeBasinmulti}
}
\quad 
\subfloat[Fitness vs. Basin of attraction size]{
  \includegraphics[scale=0.2]{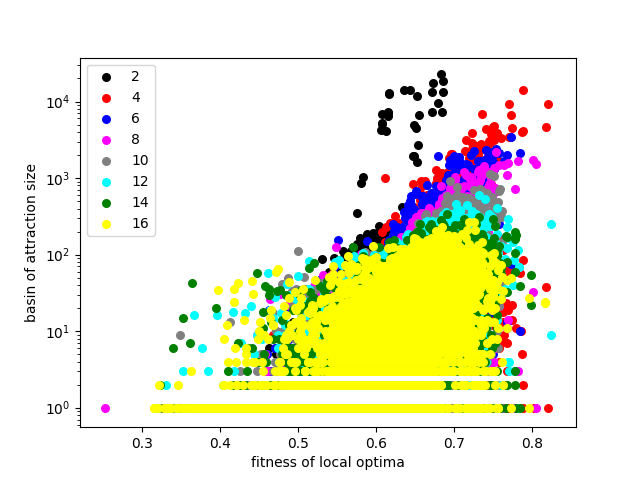}
  \label{fig:LodegreeBasinmulti}
}
\caption{Local metrics for bit-flip, swap\mar{, and multi-layer for} each value of $K$. \mar{Figures a, d and g are in log-log scale; Figures b, e, h, c, f and i are in a semi-logy scale.}}
\label{fig:local-metrics}
\end{figure*}

Complex networks are also characterized by degree distributions
\cite{antoniou2008statistical}. Therefore, in this paper we show the cumulative degree distribution besides scatter plots presenting the correlations between size of basin of attraction, degree and fitness.
Herein, we consider the strength distribution (instead of degree distribution) as weighted networks are addressed.

Comparing all local metrics in Figure \ref{fig:local-metrics}, we notice that, differently from the descriptive statistics and global metrics,  bit-flip predominates most of time, although the multi-layer network inherits   behaviors  from both layers in some particular cases.

Figures \ref{fig:degreeDistbitflip}, \ref{fig:degreeDistswap} and \ref{fig:degreeDistmulti}  show the cumulative strength distribution, \mar{in log-log scale,} considering a particular landscape for bit-flip, swap and multi-layer networks, respectively. In each figure,  one curve is shown for each value of $K$. 
The cumulative strength distribution function represents the probability $P(st=c)$ that a randomly chosen node has a strength larger than or equal to $c$.
For all the $K$ values, the cumulative strength distributions decay slowly for small 
strength values, with a faster dropping rate for high values of strength. Moreover, for bit-flip, the curves for distinct $K$ values differentiate from each other more than for swap and multi-layer. This behavior is not inherit by multi-layer since its curves seem more impacted by the swap LON. 
The curves indicate that the most part of the nodes have low strength connections, while few nodes present connections with significantly higher values of strength.
This behaviour can be observed for the bit-flip, swap and the multi-layer networks. 

Figures \ref{fig:degreeBasinbitflip}, \ref{fig:degreeBasinswap} and \ref{fig:degreeBasinmulti} present the correlation between the strength and the size of basins of attraction \mar{in a semi-logy scale}. Analyzing the bit-flip landscape, we observe that their basin sizes are greater than the swap ones showing a slightly positive correlation with strength for some values of $K$. However this behavior seems to be different in swap, that presents a high positive correlation for any value of $K$, corroborating with the stable results presented in Table \ref{tab:statistics}. Moreover, we notice that in this case, bit-flip  influences much more the multi-layer than swap.

From Figures~\ref{fig:LOdegreeBasinbitflip}, \ref{fig:LodegreeBasinswap} and \ref{fig:LodegreeBasinmulti}, which present the correlation between the fitness and the basin of attraction size \mar{in a semi-logy scale}, a low correlation between these metrics for bit-flip while a non-linear correlation, tending to lie near a smooth curve, can be observed for swap. In this case multi-layer seems to mix the behavior of both operators, in spite of reducing the correlation levels observed for both. In the case of swap, we observe a slightly better behavior since positive correlation between fitness and sizes of basins of attraction would be interesting as lots of solutions would be eventually attracted to good local optima or even the global optimum.


\ignore{
\subsection{Multi-layer metrics}

Besides, we also present some complex metrics specifically for multi-layer networks in Table \ref{tab:multi-metrics}.

\begin{table}[h!]
\small
\centering
\caption{Multi-layer Metrics}
\label{tab:multi-metrics}
\begin{tabular}{llllllll}
\hline
\toprule
\textbf{K}& 
\textbf{$dx$} & 
\textbf{$di$} & 
\textbf{$fh$}  & 
\textbf{$eg$}\\ \hline
$ 	2	$&$		$&$		$&$		$&$		$\\
$	4	$&$		$&$		$&$		$&$		$\\
$	6	$&$		$&$		$&$		$&$		$\\
$	8	$&$		$&$		$&$		$&$		$\\
$	10	$&$		$&$		$&$		$&$		$\\
$	12	$&$		$&$		$&$		$&$		$\\
$	14	$&$		$&$		$&$		$&$		$\\
$	16	$&$		$&$		$&$		$&$		$\\
\hline										
\end{tabular}
\end{table}
}

%% file: sect5.tex
\section{Conclusions and future directions}
\label{sec:conclusion}

Local Optima Networks (LONs) are graph models used in fitness landscape analysis for understanding the search difficulty, studying metaheuristics and extracting representative characteristics from problem instances. 
In this paper we analyzed LON features extracted from three different landscapes all of built from NK-landscape instances. The first landscape uses the bit-flip, the second one uses  1-swap moves and the third was obtained by combining both operators in a two-layer network. In this way we proposed a new model called multi-layer local optimal network (MLLON). In addition, formal concepts of MLLONs have been presented, including important aspects of the methodology to explore the model and extract FLA metrics from it.



The MLLON analysis showed how properties are shared from both network layers. Depending on the network metric considered, the swap operator introduces properties which are invariant with $K$. This is the case for global metrics in general. However, the bit-flip operator introduces shortcuts evidenced by high clustering coefficients and short paths to the global optimum, giving characteristics of small-world network models.


In the future, we intend to study how the proposed multi-layer model can lead to making specific design decision in the context of VNS and other metaheuristics alike. To achieve this, we aim at using more local search operators and efficiently connecting basins of attraction belonging to different landscapes. Particularly, considering the edges between basins with different local optima (which can be accomplished by setting $P_{co}^{\neq}>0$). In addition, we plan to explore other characteristics of the multi-layer networks that we did not consider in this paper, \mar{as neutrality}, and exploring other forms of calculating metrics for multi-layer networks (e.g. by layer aggregation instead of flattening).

%% file: arXiv.bbl
\begin{thebibliography}{10}

\bibitem{LonContProbGecco2019}
Jason Adair, Gabriela Ochoa, and Katherine~M. Malan.
\newblock Local optima networks for continuous fitness landscapes.
\newblock In {\em Proceedings of the Genetic and Evolutionary Computation
  Conference Companion}, page 1407–1414, New York, NY, USA, 2019. Association
  for Computing Machinery.

\bibitem{antoniou2008statistical}
Ioannis~E Antoniou and ET~Tsompa.
\newblock Statistical analysis of weighted networks.
\newblock {\em Discrete dynamics in Nature and Society}, 2008, 2008.

\bibitem{LonContProb2017}
Andrew~J. Ballard, Ritankar Das, Stefano Martiniani, Dhagash Mehta, Levent
  Sagun, Jacob~D. Stevenson, and David~J. Wales.
\newblock Energy landscapes for machine learning.
\newblock {\em Phys. Chem. Chem. Phys.}, 19:12585--12603, 2017.

\bibitem{barrat2004architecture}
Alain Barrat, Marc Barthelemy, Romualdo Pastor-Satorras, and Alessandro
  Vespignani.
\newblock The architecture of complex weighted networks.
\newblock {\em Proceedings of the national academy of sciences},
  101(11):3747--3752, 2004.

\bibitem{barthelemy2005characterization}
Marc Barth{\'e}lemy, Alain Barrat, Romualdo Pastor-Satorras, and Alessandro
  Vespignani.
\newblock Characterization and modeling of weighted networks.
\newblock {\em Physica a: Statistical mechanics and its applications},
  346(1-2):34--43, 2005.

\bibitem{chicano2012local}
Francisco Chicano, Fabio Daolio, Gabriela Ochoa, S{\'e}bastien V{\'e}rel, Marco
  Tomassini, and Enrique Alba.
\newblock Local optima networks, landscape autocorrelation and heuristic search
  performance.
\newblock {\em Parallel Problem Solving from Nature (PPSN)}, pages 337--347,
  2012.

\bibitem{chicano2017optimizing}
Francisco Chicano, Darrell Whitley, Gabriela Ochoa, and Renato Tin{\'o}s.
\newblock {Optimizing one million variable NK landscapes by hybridizing
  deterministic recombination and local search}.
\newblock In {\em Genetic and Evolutionary Computation Conference (GECCO)},
  pages 753--760. ACM, 2017.

\bibitem{costa2007characterization}
L~da~F Costa, Francisco~A Rodrigues, Gonzalo Travieso, and Paulino~Ribeiro
  Villas~Boas.
\newblock Characterization of complex networks: A survey of measurements.
\newblock {\em Advances in physics}, 56(1):167--242, 2007.

\bibitem{daolio2010local}
Fabio Daolio, S{\'e}bastien Verel, Gabriela Ochoa, and Marco Tomassini.
\newblock Local optima networks of the quadratic assignment problem.
\newblock In {\em IEEE Congress on Evolutionary Computation (CEC)}, pages 1--8.
  IEEE, 2010.

\bibitem{daolio2012local}
Fabio Daolio, S{\'e}bastien Verel, Gabriela Ochoa, and Marco Tomassini.
\newblock Local optima networks and the performance of iterated local search.
\newblock In {\em Genetic and Evolutionary Computation Conference}, GECCO,
  pages 369--376. ACM, 2012.

\bibitem{daolio2013local}
Fabio Daolio, S{\'e}bastien Verel, Gabriela Ochoa, and Marco Tomassini.
\newblock Local optima networks of the permutation flow-shop problem.
\newblock In {\em International Conference on Artificial Evolution (Evolution
  Artificielle)}, pages 41--52. Springer, 2013.

\bibitem{de2013mathematical}
Manlio De~Domenico, Albert Sol{\'e}-Ribalta, Emanuele Cozzo, Mikko Kivel{\"a},
  Yamir Moreno, Mason~A Porter, Sergio G{\'o}mez, and Alex Arenas.
\newblock Mathematical formulation of multilayer networks.
\newblock {\em Physical Review X}, 3(4):041022, 2013.

\bibitem{dickison2016multilayer}
Mark~E Dickison, Matteo Magnani, and Luca Rossi.
\newblock {\em Multilayer social networks}.
\newblock Cambridge University Press, 2016.

\bibitem{hagberg2008exploring}
Aric Hagberg, Pieter Swart, and Daniel S~Chult.
\newblock Exploring network structure, dynamics, and function using networkx.
\newblock Technical report, Los Alamos National Lab.(LANL), Los Alamos, NM
  (United States), 2008.

\bibitem{VNS_survey2016}
P.~Hansen, N.~Mladenovic, R.~Todosijevic, and S.~Hanafi.
\newblock Variable neighborhood search: basics and variants.
\newblock {\em EURO Journal on Computational Optimization}, 5, 08 2016.

\bibitem{hernando2017local}
Leticia Hernando, Fabio Daolio, Nadarajen Veerapen, and Gabriela Ochoa.
\newblock Local optima networks of the permutation flowshop scheduling problem:
  Makespan vs. total flow time.
\newblock In {\em IEEE Congress on Evolutionary Computation (CEC)}, pages
  1964--1971. IEEE, 2017.

\bibitem{jakobovic2019characterisation}
Domagoj Jakobovic, Stjepan Picek, Marcella~SR Martins, and Markus Wagner.
\newblock A characterisation of s-box fitness landscapes in cryptography.
\newblock In {\em Proceedings of the Genetic and Evolutionary Computation
  Conference}, pages 285--293, 2019.

\bibitem{kauffman1993origins}
Stuart~A Kauffman.
\newblock {\em The origins of order: Self-organization and selection in
  evolution}.
\newblock Oxford University Press, USA, 1993.

\bibitem{kivela2014multilayer}
Mikko Kivel{\"a}, Alex Arenas, Marc Barthelemy, James~P Gleeson, Yamir Moreno,
  and Mason~A Porter.
\newblock Multilayer networks.
\newblock {\em Journal of complex networks}, 2(3):203--271, 2014.

\bibitem{newman2002assortative}
Mark~EJ Newman.
\newblock Assortative mixing in networks.
\newblock {\em Physical review letters}, 89(20):208701, 2002.

\bibitem{ochoa2008study}
Gabriela Ochoa, Marco Tomassini, Seb{\'a}stien V{\'e}rel, and Christian
  Darabos.
\newblock {A study of NK landscapes' basins and local optima networks}.
\newblock In {\em Genetic and Evolutionary Computation Conference (GECCO)},
  pages 555--562. ACM, 2008.

\bibitem{ochoa2017mapping}
Gabriela Ochoa and Nadarajen Veerapen.
\newblock Mapping the global structure of tsp fitness landscapes.
\newblock {\em Journal of Heuristics}, 24:265–--294, 2018.

\bibitem{ochoa2014local}
Gabriela Ochoa, S{\'e}bastien Verel, Fabio Daolio, and Marco Tomassini.
\newblock Local optima networks: A new model of combinatorial fitness
  landscapes.
\newblock In {\em Recent Advances in the Theory and Application of Fitness
  Landscapes}, pages 233--262. Springer, 2014.

\bibitem{ochoa2010first}
Gabriela Ochoa, S{\'e}bastien Verel, and Marco Tomassini.
\newblock First-improvement vs. best-improvement local optima networks of nk
  landscapes.
\newblock {\em Parallel Problem Solving from Nature, PPSN XI}, pages 104--113,
  2010.

\bibitem{Skrlj2019}
Blaz Skrlj, Jan Kralj, and Nada Lavrac.
\newblock Py3plex toolkit for visualization and analysis of multilayer
  networks.
\newblock {\em Applied Network Science}, 4(1):94, 2019.

\bibitem{thomson2018fractal}
Sarah Thomson, S{\'e}bastien Verel, Gabriela Ochoa, Nadarajen Veerapen, and
  Paul McMenemy.
\newblock On the fractal nature of local optima networks.
\newblock In {\em EvoCOP 2018-The 18th European Conference on Evolutionary
  Computation in Combinatorial Optimisation}. Springer, 2018.

\bibitem{tomassini2008complex}
Marco Tomassini, S{\'e}bastien Verel, and Gabriela Ochoa.
\newblock {Complex-network analysis of combinatorial spaces: The NK landscape
  case}.
\newblock {\em Physical Review E}, 78(6):066114, 2008.

\bibitem{verel2011localESCAPE}
S{\'e}bastien V{\'e}rel, Fabio Daolio, Gabriela Ochoa, and Marco Tomassini.
\newblock Local optima networks with escape edges.
\newblock In {\em Artificial Evolution}, pages 49--60. Springer, 2011.

\bibitem{verel2018sampling}
S{\'e}bastien Verel, Fabio Daolio, Gabriela Ochoa, and Marco Tomassini.
\newblock Sampling local optima networks of large combinatorial search spaces:
  the qap case.
\newblock In {\em International Conference on Parallel Problem Solving from
  Nature}, pages 257--268. Springer, 2018.

\bibitem{verel2008connectivity}
S{\'e}bastien Verel, Gabriela Ochoa, and Marco Tomassini.
\newblock The connectivity of nk landscapes' basins: A network analysis.
\newblock In {\em Artificial Life XI: 11th International Conference on the
  Simulation and Synthesis of Living Systems}, pages 648--655, 2008.

\bibitem{verel2011localNK}
S{\'e}bastien Verel, Gabriela Ochoa, and Marco Tomassini.
\newblock {Local optima networks of NK landscapes with neutrality}.
\newblock {\em IEEE Transactions on Evolutionary Computation}, 15(6):783--797,
  2011.

\bibitem{Yafrani2018FLA}
Mohamed~El Yafrani, Marcella S.~R. Martins, Mehdi~El Krari, Markus Wagner,
  Myriam R. B.~S. Delgado, Bela\"{\i}d Ahiod, and Ricardo L\"{u}ders.
\newblock A fitness landscape analysis of the travelling thief problem.
\newblock In {\em Genetic and Evolutionary Computation Conference (GECCO)},
  pages 277--284. ACM, 2018.

\end{thebibliography}
